\documentclass{article}
\usepackage{nac_preprint}

\usepackage{listings}
\usepackage{times}
\usepackage{epsfig}
\usepackage{graphicx}
\usepackage{amsmath}
\usepackage{amssymb}
\usepackage{xcolor}
\usepackage{caption}
\usepackage{enumitem}

\usepackage[symbol]{footmisc}

\usepackage[numbers,sort]{natbib} 

\DeclareCaptionType{exam}[Example][List of Examples]
\usepackage[pagebackref=true,breaklinks=true,letterpaper=true,colorlinks,bookmarks=false]{hyperref}

\begin{document}

\title{Predicted Embedding Power Regression for Large-Scale \\Out-of-Distribution Detection} 

\author{Hong Yang, William Gebhardt, Alexander G. Ororbia\thanks{Equal advising.}, and Travis Desell\footnotemark[1]\\
Rochester Institute of Technology\\
1 Lomb Memorial Dr, Rochester, NY 14623\\
{\tt\small [hy3134
wdg1351, agovcs, tjdvse]@rit.edu}
}

\maketitle

\begin{abstract}
Out-of-distribution (OOD) inputs can compromise the performance and safety of real world machine learning systems. While many methods exist for OOD detection and work well on small scale datasets with lower resolution and few classes, few methods have been developed for large-scale OOD detection. Existing large-scale methods generally depend on maximum classification probability, such as the state-of-the-art grouped softmax method. In this work, we develop a novel approach that calculates the probability of the predicted class label based on label distributions learned during the training process. Our method performs better than current state-of-the-art methods with only a negligible increase in compute cost. We evaluate our method against contemporary methods across $14$ datasets and achieve a statistically significant improvement with respect to AUROC ($84.2$ vs $82.4$) and AUPR ($96.2$ vs $93.7$).  
\end{abstract} %

\section{Introduction}
\label{sec:intro}

Out of distribution (OOD) detection is a critical tool for the development of safe and reliable autonomous and semi-autonomous machine learning (ML) systems. Identification of anomalous inputs allows intelligent systems to initiate a conservative fallback policy or to  defer to human judgment, reducing the risks to all stakeholders. Hendrycks \cite{hendrycks2021unsolved} noted that OOD detection is important for safety critical systems, such as self-driving cars and detecting novel microorganisms. As a result, a plethora of literature has emerged over the years for addressing the problem of OOD detection \cite{hendrycks2016baseline, hendrycks2018deep, bevandic2018discriminative, lakshminarayanan2017simple, liang2017enhancing}. However, these efforts focus on small and lower resolution datasets. Although, recent work has attempted to address the issue of large scale OOD detection \cite{hendrycks2019scaling, huang2021mos} on large datasets at high resolution \cite{deng2009imagenet, ridnik2021imagenet}. 

Emerging safety critical systems operate on a much larger dataset at much higher resolutions, the real world. Unlike small datasets, real world data often contains hundreds or thousands of classes at a very high resolution. Use cases for OOD detection range from autonomous driving \cite{bogdoll2022anomaly} to medical imaging \cite{maartensson2020reliability}, applications that contain a significant number of potential classes. However, the reliability of the typical baseline OOD detection method \cite{hendrycks2016baseline} decreases rapidly as the number of classes increases, resulting in a change from $17.3$\% false positive rate at $95$\% true positive rate (FPR95) with $50$ classes to a $76.9$\% FPR95 when using $1000$ classes \cite{huang2021mos}. 

Our research is motivated by the need to improve OOD detection methods for use in  safety critical applications.  We focus on the theoretical concept of Bayesian conditional probability for improving the decision boundary between in-distribution and OOD data. If we consider softmax classification as estimating the class $y$ given data $x$ as $P(y|x)$, our method considers modeling $P(x|\hat{y})$, the probability of the data given the predicted class itself. Doing so will allow us to further consider the marginal probability of $P(\hat{y})$ existing within our in-distribution data, leading us to develop an efficient and effective method for scoring OOD data. 

We estimate $P(\hat{y})$ by leveraging the behavioural characteristics of the exponential linear unit (ELU) \cite{clevert2015fast} activation function and the process of batch normalization \cite{ioffe2015batch}. Notably, we find that combining the ELU function with batch normalization results in sparse, large values for an embedding $z$ prior to the final classification layer for in distribution data. A large expected embedding value $\mathbb{E}(z|\hat{y})$ functions as a proxy for $P(\hat{y})$, such that larger expected embedding values indicate a $\hat{y}$ that has been observed repeatedly during training, see Figure \ref{fig:peprhist1}.

We extensively evaluate our approach on models trained with the Imagenet $1$k dataset \cite{deng2009imagenet}, leveraging the state-of-the-art pre-trained BiT-S models \cite{kolesnikov2020big} as our pre-trained model backbone. We significantly reduce required computation and memory usage 
by pre-computing our backbone outputs and demonstrate that our efficient approach 
successfully reproduces the results from \cite{huang2021mos}, the state-of-the-art large-scale OOD detection task on Imagenet $1$K. Compared to the previous best method \cite{huang2021mos}, we demonstrate that our method scores higher in terms of area under the receiver operating characteristic (AUROC) curve ($84.2$ versus $82.4$)  and area under the precision-recall (AUPR) curve ($96.2$ versus $93.7$) on a larger, more diverse set of benchmarks. The results of this paper mark an important step towards leveraging the conditional probability of intermediate outputs for OOD detection. Below, we summarize this study's \textbf{key results and contributions:} 
\begin{itemize}[noitemsep,nolistsep]
    \item We propose a novel conditional probability-based scoring method, the Predicted Embedding Power Regression (PEPR) method and the Combined PEPR (C-PEPR) method, that performs better than current state-of-the-art by a statistically significant margin. 
    \item We introduce a high variance OOD detection method that benefits from ensembling. Contemporary methods tend to exhibit extreme stability across runs for all measured performance metrics. In contrast, our approach's stochasticity results in ensembling benefits. 
    \item We reproduce contemporary methods across a larger group of benchmarks and demonstrate that our computational method is more energy and compute efficient, reducing the training step time by nearly $80$\%. Unlike previous papers, our experiments measure the standard deviation across multiple runs and establish a statistically significant improvement over the state of the art.  
\end{itemize}

\section{Related Work}
\label{sec:related_work}

\noindent
\textbf{Multi-class OOD Detection with Pre-trained Models:} 
A common baseline for OOD was historically established by Hendrycks and Gimpel \cite{hendrycks2016baseline} which specifically used the maximum softmax probability. Further efforts have attempted to improve the OOD estimation by using the ODIN score \cite{liang2017enhancing}, deep ensembles \cite{lakshminarayanan2017simple}, a Mahalanobis distance-based confidence score \cite{lee2018simple}, Energy score \cite{liu2020energy}, as well as the Minimum Other Score \cite{huang2021mos}. Note that these methods do not use any OOD data for fine tuning or training. 

\noindent
\textbf{Multi-class OOD Detection with Model Fine-tuning:} 
An alternative research direction is to leverage additional data from outside of the distribution in order to regularize the model \cite{bevandic2018discriminative}, \cite{geifman2019selectivenet}, \cite{malinin2018predictive}. In this setup, additional/auxiliary data may be realistic images \cite{hendrycks2018deep} or synthetic images generated by generative adversarial networks (GANs) \cite{lee2017training}. These extra images are used in various ways, such as to regularize the probabilities back to a uniform distribution \cite{lee2017training} or as ``background'' class samples \cite{mohseni2020self}. Note that constructing additional OOD data assumes a distribution for such data points. This means that the additional OOD data may not be similar to other OOD data samples that the model would encounter in the wild. 

\noindent
\textbf{Large Scale Multi-class OOD Detection:} 
This line of work focuses on datasets with a large number of classes, most commonly found in Imagenet $1$k. \cite{roady2019out} utilized half of Imagenet $1$k as in-distribution data and the other half as out-of-distribution data. They also used the Places-$434$ dataset and evaluated a plethora of different approaches, including KL matching and MSP. \cite{huang2021mos} introduced a grouped cross-entropy to generate an implicit inter-group background class, which was demonstrated to improve system performance. \cite{hendrycks2019scaling} further increased the number of classes by using data from Imagenet $21$k.

\noindent
\textbf{Hierarchical Classification:} 
Hierarchical structure can provide additional label information, which can facilitate efficient inference \cite{deng2011fast}, improved generalization accuracy \cite{deng2014large}, and better object detection \cite{redmon2017yolo9000}. Some efforts have made use of a label tree structure when a taxonomy is unavailable \cite{deng2011fast,deng2014large}. Many studies explore the benefits and importance of basic hierarchical structures for various classification tasks \cite{warde2014self,hinton2015distilling,gross2017hard}. 

\noindent
\textbf{Bayesian Randomised MAP Sampling:} 
This line of related work exploits the fact that adding a regularisation term to a loss function returns a maximum a posteriori (MAP) parameter estimate, i.e., a point estimate of the Bayesian posterior. Repeating this calculation produces a distribution of MAP solutions that mimicks that of the true posterior. This allows for efficient sampling of high-dimensional posteriors \cite{chen2012ensemble}. Some methods allow for sampling of the posterior but fail to recover the true posterior itself \cite{lu2017ensemble}. In contrast, other methods require significant computational resources for recovering the true posterior \cite{bardsley2014randomize}. Work done in \cite{pearce2020uncertainty} provided a suitable compromise with respect to the accuracy of the posterior and an increase in computational cost. 

\section{Methodology} 
\label{sec:methodology}

\subsection{Preliminaries}
\label{sec:prelim}

We consider a training dataset drawn i.i.d. from the in-distribution $P_X$, with label space $Y =
{1, 2,  \hdots , C}$. For the OOD detection problem, we train a classifier $F(x)$ on the in-distribution $P_X$, and evaluate it on samples that are drawn from a different (outside) distribution $Q_X$ . An OOD detector $G(x)$ is a binary classifier defined as:
\begin{align}
    G(\mathbf{x})= 
    \begin{cases} 
    1 & \text { if } S(\mathbf{x}) \geq \gamma \quad \text{// ``in''} \\ 
    0 & \text { if } S(\mathbf{x})<\gamma \quad \text{// ``out''}
    \end{cases}     
\end{align}
where $S(x)$ is a scoring function and $\gamma$ is a threshold, determined by the target practical application of the OOD detector (different applications would have different precision and recall targets).  

\subsection{KL Matching and Class Similarity}
KL matching \cite{hendrycks2019scaling} provides a useful intuition for solving large scale OOD detection. As the number of classes/categories increases, the similarity between classes may also increase, leading to reduced confidence in the classifier's predictions. This means that the performance of the Maximum Softmax Probability (MSP) \cite{hendrycks2016baseline} method for OOD degrades with a greater number of classes. KL matching attempts to solve the class size problem by generating expected probability distributions around each category, a.k.a. posterior distribution templates, expecting in-distribution image patterns to more closely match the posterior distribution templates. In effect, it measures $P(\hat{y})$, the probability of the class existing in the in-distribution dataset. 

However, KL matching has a major disadvantage compared to other methods; it needs to store the posterior distribution templates and use them for comparison. When performing OOD detection, each image's predicted class distribution must be compared against each distribution template in order to find the minimum KL divergence value. This means that the computational cost scales with the square of the number of classes. Although this is not problematic with $1000$ classes, it quickly becomes prohibitive when there tens of thousands of classes or more.

\begin{figure}
    \centering
    \includegraphics[width=11cm]{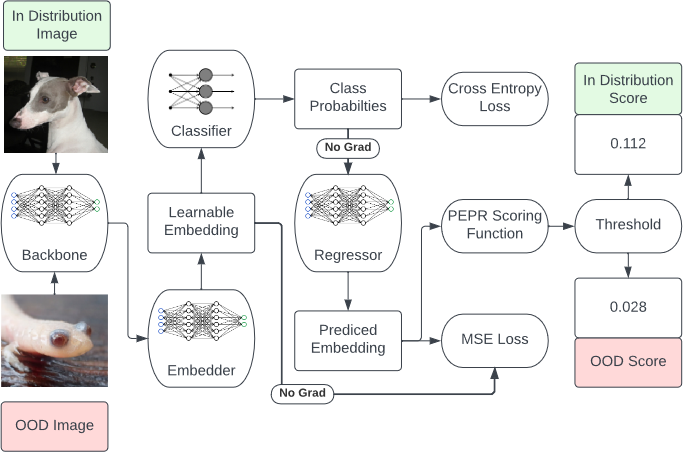}
    \caption{Overview of the PEPR model. The process consists of the following steps: 
    \textbf{(1)} Learn to classify training images via the learnable embedding, 
    \textbf{(2)} Estimate the embedding values conditioned on the predicted class probabilities, and 
    \textbf{(3)} Define a threshold and use PEPR to calculate the score. No gradient flows between parts \textbf{(1)} and \textbf{(2)} and no OOD images are used during training. Note that the classifier can be any fully connected layer, depending on the number of classes. Definitions for the embedder and regressor can be seen in listings \ref{embedder} and \ref{regressor}}.
    \label{fig:example}
\end{figure}

\subsection{Predicted Embedding Power Regression}
\label{sec:pepr_score}

\begin{figure}
    \centering
    \includegraphics[width=9cm]{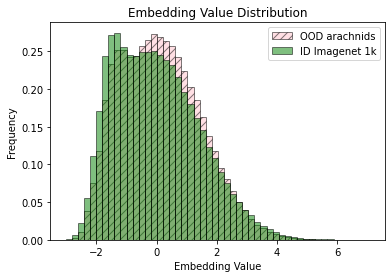}
    \caption{Distribution of embedding $z$ for the OOD dataset arachnids compared to the in distribution dataset of Imagenet 1K. There is a noticeable tail in the distribution of the in-distribution dataset. These distributions overlap significantly. Frequency is such that the area sums to one.}
    \label{fig:embhist1}
\end{figure}

\begin{figure}
    \centering
    \includegraphics[width=9cm]{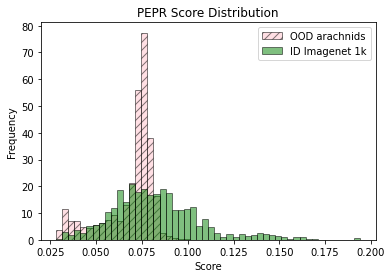}
    \caption{The distribution of $S_{\text{PEPR}}(x)$ for the OOD data \textit{arachnids} compared to the Imagenet $1$K in-distribution data. The regressor $\hat{z} = R(\hat{y})$ predicts significantly higher values for $\hat{z}$ for in-distribution patterns when compared to OOD ones. }
    \label{fig:peprhist1}
\end{figure}

In this work, we use the intuition behind KL matching to formulate a new conditional probability test that does not rely on an array of posterior distribution templates.  Concretely, we consider the positive expectation of a model's selected intermediate layer outputs conditioned on the predicted softmax probability distribution. Note that this test is crucially built on two key theoretical notions. 

First, if a selected intermediate layer uses a linear rectifier based (ReLU-like) activation function followed by a batch normalization (batch norm) operation, that layer's output can be viewed as an embedding $z$. A negative value in this layer's activity represents the absence of a feature and a positive value represents its presence. We observe empirically that such a layer, which precedes the final softmax classification output layer, exhibits a tail-heavy behaviour, with a large proportion of values below zero (the batch norm mean) and a minority of values significantly above zero. Figure \ref{fig:embhist1} displays the distribution of embedding values of such a layer after model training. This means that, for any given image, we would expect to observe a minority consisting of very large positive values and a majority of consisting of negative values in $z$. 

Second, an additional model, which we will refer to as the regressor $\hat{z} = R(\hat{y})$, will learn to predict the expected value of the embedding layer, i.e,  $\mathbb{E}(z | \hat{y})$, where the predicted class distribution is $F(x) = \hat{y}$. When $R(\hat{y})$ is trained using a mean squared error loss, the regressor should learn to predict mostly negative values and a few large positive values. For class distributions that are not present in the training data, we would expect $R(\hat{y})$ to predict values close to zero, which is the true batch normalized mean for the embedding unconditional of the class, i.e., $\mathbb{E}(z) = 0$. 

Combining the above two notions together yields our proposed OOD detection method, i.e., the predicted embedding power regression (PEPR) model. By using an intermediate layer between the model backbone and the softmax classification layer, PEPR can then focus on modeling a batch normalized embedding. We then train a nonlinear regression model to estimate the embedding values based on the softmax classification distribution. Note that we expect that the average of the squared positive expected embedding values to be higher for in-distribution data than for OOD data. In effect, we recover an estimate of $P(\hat{y})$ via the magnitude of $\hat{z}$, which is learned via regressor $R(\hat{y})$ from training data patterns. Note that our method only adds a small amount of computational overhead. The actual score computed using our regressor is formally:
\begin{align}
    S_{\text{PEPR}}(x) = \frac{1}{n} \sum_{i=1}^n \Big( \text{ReLU}\big( R(\hat{y})_i \big) \Big)^2 
    \label{pepr}
\end{align}
where $n$ is the dimensionality of the embedding. An empirical sample of the PEPR score distribution is presented in Figure \ref{fig:peprhist1}, which corroborates our hypothesis. 

Finally, we further improve the PEPR model by also utilizing the actual embedding values. Specifically, we compute the actual embedding in-distribution score by calculating the mean of the squared embedding values, which we call the embedding power (EPOW). We then add this score, weighted by the coefficient parameter $\psi$, to the PEPR score and refer to the final score as C-PEPR. Formally, this score is calculated as follows:
\begin{align}
    S_{\text{C-PEPR}}(x) = S_{\text{PEPR}}(x) + \psi \frac{1}{n} \sum_{i=1}^n  (z_i)^2 \label{cpepr}
\end{align}
Note that, in this work, we set the factor $\psi = 0.01$.

\subsubsection{Stability}
\label{sec:stability}

Our empirical results indicated that the PEPR and C-PEPR methods have a higher standard deviation than contemporary methods. Our methods tend to have a standard deviation of $\approx$ 0.9 AUROC, while MSP \cite{hendrycks2016baseline} and Minimum Other Score (MOS) \cite{huang2021mos} have a standard deviation of less than $0.1$ AUROC. We decided to evaluate a 10 regressor ensemble of C-PEPR (C-PEPR-10) and a 10 classifier ensemble of MOS (MOS-10) to investigate the stability and AUROC improvements of ensembling. 

\subsubsection{Grouped Labels}
\label{sec:grouped_labels}

In this paper, we train CPEPR and PEPR using the grouped softmax approach presented in \cite{huang2021mos}. Experimental results indicated that PEPR did not work quite as well with the standard softmax setup (average AUROC 78.4 vs 84.2). Detailed results for PEPR using standard softmax will be provided in the supplementary material. We believe that this issue may be caused by the high level of sparsity in the softmax probability values. Another possible explanation for this issue is that the learned embeddings are better when using the grouped softmax approach. Either way, grouped softmax PEPR achieves state of the art performance. 

\subsubsection{Bayesian Ensembling}
\label{sec:bayes_ensemble}

Notably, we integrated the anchored ensembling approach presented in  \cite{pearce2020uncertainty}. This scheme functions almost identically to L2 regularization of model weights, except that the regularization target is the random initial weight values rather than zero (as done in traditional parameter regularization, which assumes a zero-mean Gaussian prior over parametres). By taking an ensemble of networks regularized in this manner, Pearce \cite{pearce2020uncertainty} demonstrated that one can emulate the desired behaviour of a Bayesian neural network (without the prohibitive cost). While initially we intended to utilize Bayesian behaviour for OOD, PEPR and C-PEPR often worked quite well with an ensemble size of one. However, whether or not the regressor $R(\hat{y})$ is ensembled, our scoring models do require anchored regularization in order to achieve the best results presented as presented in Section \ref{sec:results}. 

Note that the anchored regularization scheme is only applied to our regressor $R(\hat{y})$ and not to the model classifier, embedding component, or model backbone. Equation \ref{anch} describes the anchored mean squared error loss for the $j$th (regressor) model for batches of size $N$ with weights $\theta_j$ and initial weight values ${\theta}_{\text{anc}, j}$. 
\begin{align}
\operatorname{Loss}_j =\frac{1}{N}\left\|\mathbf{z}-\hat{\mathbf{z}}_j\right\|_2^2+\frac{1}{N}\left\|\gamma \cdot\left(\boldsymbol{\theta}_j-\boldsymbol{\theta}_{\text{anc}, j}\right)\right\|_2^2    \label{anch}
\end{align}
We note that $\hat{z}_j$ is the predicted value of $z$ from the $j$th regressor model $R_j(\hat{y})$. We treat $\gamma$ as a hyper-parameter and set it to $0.03$ for all experiments unless otherwise specified.

\section{Experiments} 
\label{sec:results}

We provide a Github repository to fully replicate the experiments. We also provide Colab notebooks to allow users without access to local compute resources to run the experiments for themselves.

\subsection{Datasets}
\label{sec:data}

\subsubsection{In-Distribution Dataset}
We use Imagenet $1$k as our in-distribution dataset \cite{deng2009imagenet}. This dataset has been used for large-scale OOD experiments such as those conducted by \cite{huang2021mos} and \cite{hendrycks2019scaling}, which allows us to properly compare methods.

\begin{figure}
    \centering
    \includegraphics[width=9cm]{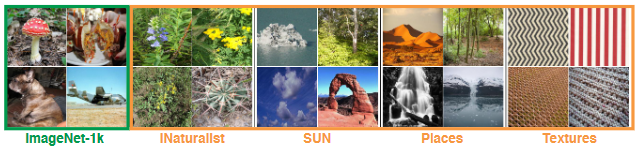}
    \caption{Samples of in-distribution/OOD data (as in \cite{huang2021mos}).}
    \label{fig:samples_huang}
\end{figure}

\begin{figure}
    \centering
    \includegraphics[width=8.25cm]{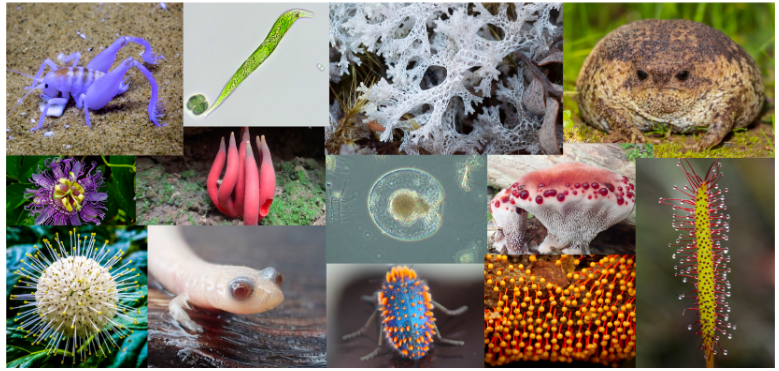}
    \caption{Samples from OOD datasets, as in \cite{hendrycks2019scaling}. Images  extracted from the all species datasets.}
    \label{fig:samples_hendrycks}
\end{figure}

\subsubsection{Out-of-Distribution Datasets}
We used two sets of out-of-distribution (OOD) datasets. First is curated version of Textures, SUN, Places, and iNaturalist benchmarks, presented in \cite{huang2021mos}. Second, is the the Anomalous Species Dataset, presented in \cite{hendrycks2019scaling}. 

\noindent 
\textbf{iNaturalist:} iNaturalist \cite{van2018inaturalist} contains $859,000$ images of $5,000$ species of plants and animals. \cite{huang2021mos} manually selected $110$ plant
classes not present in ImageNet-$1$k and randomly sampled $10,000$ images from these $110$ classes. All images were resized to have a maximum dimension of $800$ pixels.

\noindent
\textbf{SUN:} SUN \cite{xiao2010sun} is a scene database of $397$ categories across $130,519$ images with sizes larger than $200 \times 200$. We used the curated version presented by \cite{huang2021mos}, which randomly selected $10,000$ images from $50$ concepts not in ImageNet $1$k.

\noindent
\textbf{Textures:} Textures \cite{cimpoi2014describing} consists of $5,640$ images of textural patterns, with sizes ranging between $300 \times 300$ and $640 \times 640$. Huang \cite{huang2021mos} uses the full dataset.

\noindent 
\textbf{Places:} Places365 \cite{zhou2017places} is a scene dataset that is similar to SUN. \cite{huang2021mos} resized all of the images to have a minimum dimension of $512$. They \cite{huang2021mos} randomly selected $10,000$ images from $50$ concepts not in ImageNet $1$k. It is unclear why \cite{huang2021mos} curated Places365 while \cite{hendrycks2019scaling} used Places365 without curation as the OOD set for Imagenet $1$k.

\noindent 
\textbf{Anomalous Species Dataset:} The Species dataset \cite{hendrycks2019scaling} contains $700,000$ species from the \cite{van2018inaturalist} dataset that do not overlap with Imagenet $21$k \cite{ridnik2021imagenet}. Due to the quantity of images, we limit each species group to $12,800$ images, with the exception of the micro-organisms group, where we use only $1,408$ images due to the lack of images. 

\subsection{PEPR Embedder and Regressor}

Unlike other OOD detection methods, PEPR requires a learnable intermediate layer to generate embeddings $z$ and a regression model to estimate the embedding conditioned on the predicted softmax probability, i.e.,  $\hat{z} = R(\hat{y})$. Unless specified otherwise, the embedding model and regressor models are feedforward neural networks (FNNs) -- the embedding FNN is specified in Listing \ref{embedder} while the regressor FNN is specified in Listing \ref{regressor}. Note that only the embedding model's outputs are used as input to the (softmax) classifier, while other methods use the backbone's outputs directly as inputs to the classifier. Furthermore, the regressor $R(\hat{y})$ gradients must not flow to the embedding or classifier FNN modules. The interaction between these components is visualized in Figure \ref{fig:example}. 

\begin{lstlisting}[caption=Embedding Model, label={embedder}]
tf.keras.Sequential([
        Dense(512, activation='elu'),
        BatchNormalization(),
        Dropout(0.1),
        Dense(512, activation='elu'),
        BatchNormalization(),
        Dropout(0.1),
        Dense(256, activation='elu'),
        BatchNormalization()])

\end{lstlisting}

\begin{lstlisting}[caption=Regression Model, label={regressor}]
tf.keras.Sequential([
    Dense(512, activation='tanh'),
    Dense(256, activation='tanh'),
    Dense(256, activation=None),
    LeakyReLU(0.1)])

\end{lstlisting}

\subsection{Experiment Setup} \label{experimentsetup}

\noindent 
\textbf{Pre-trained Backbone:} Similar to \cite{huang2021mos}, we use the Google BiT-S-R101x1 model \cite{kolesnikov2020big} with a depth of $101$ and width factor of one. Pre-trained models facilitate the extraction of high-quality features with minimal time and energy consumption. We choose to fix the backbone of our system  and only train the final layers for each method investigated. 

\noindent 
\textbf{Pre-Computed Backbone Outputs:} 
Due to the significant number of trials required for our experiments, we decided to significantly reduce energy consumption (and thus our computational carbon footprint) by pre-computing the backbone's outputs and then re-using these for downstream simulation. Specifically, we pre-computed over $12$ million backbone output vectors for Imagenet $1$k, ensuring that each image contained multiple augmented backbone outputs. On a TPUv2-8 at batch size $512$, pre-computed backbone outputs require $\approx 115$ milliseconds (ms) for one training step while computing the backbone outputs during training requires $\approx 952$ ms for one training step (due to memory limits, we must use gradient accumulation when computing backbone values). This is a $80$\% reduction in training time which results in a significant reduction in electricity usage and thus green house gas emissions. To ensure validity, we calculate the validation and OOD testing steps by calculating the backbone outputs at validation and test time, i.e., we run the model with input images during validation/testing. 

\noindent 
\textbf{Training Details:} 
All models are trained using the Adam optimizer \cite{kingma2014adam} with a step size of $0.0003$ until the $9$th epoch, where the learning rate is decreased by $40$\% (and again in the 10th epoch). We train for $10$ epochs with batches of $512$ samples and $1000$ steps per epoch, i.e., a total of $10$k steps. We decided that the BIT Hyperrule \cite{kolesnikov2020big} does not apply if we choose to freeze the backbone or use pre-computed backbone outputs. When pre-computing backbone outputs, all images are resized to $512 \times 512$ and randomly cropped to $480 \times 480$ (using a random horizontal flip). At test time, all images are resized to $480 \times 480$. At both pre-computing and test time, images are normalized as in \cite{kolesnikov2020big}. We perform all experiments using TPUv$2$-$8$s on Google Colab. 

\noindent 
\textbf{Evaluation Metrics:}  
We measure the following metrics commonly used in OOD detection: 
(1) the false positive rate of OOD examples when the true positive rate of in-distribution examples is at $95$\% (FPR95); 
(2) the area under the receiver operating characteristic curve (AUROC); and 
(3) the area under the precision-recall curve (AUPR). Note that we run each experiment $10$ times and report the mean and standard deviation of the measurements. 

We note that the KL matching method described by \cite{hendrycks2019scaling} calculated the posterior distribution templates using the in-distribution dataset labels. This can be problematic if we consider an in-distribution dataset with one image per class. In such a case, the KL divergence of each image's predicted distribution versus the distribution templates would be minimal (as they would be the same values). To address this issue, we calculate the posterior distribution templates using the training data itself. We believe this to be an accurate representation of the KL matching method as \cite{hendrycks2019scaling} noted that the in-distribution dataset labels are not necessary for the KL matching method. For the above reasons, we recommend future researchers calculate KL matching without using the in-distribution test dataset labels.

\section{Results}

\begin{table}[!t]

\centering
\begin{tabular}{|l|rrr|}

\hline
 & AUROC & AUPR & FPR95 \\
Method & Mean ±$\sigma$ & Mean ±$\sigma$ & Mean ±$\sigma$  \\
\hline
C-PEPR (ours) & \textbf{84.2} ±0.9 & \textbf{96.2} ±0.3 & 56.9 ±1.3 \\
\cline{1-4}
C-PEPR-10 (ours) & \textbf{84.6} ±0.8 & \textbf{96.3} ±0.3 & 57.0 ±1.1 \\
\cline{1-4}
EPOW & 74.9 ±0.8 & 93.4 ±0.2 & 75.7 ±1.8 \\
\cline{1-4}
KLM & 78.9 ±0.1 & 93.4 ±0.0 & 70.0 ±0.2 \\
\cline{1-4}
MLGT & 73.0 ±0.1 & 92.7 ±0.0 & 90.0 ±0.1 \\
\cline{1-4}
MOS & 82.4 ±0.1 & 93.7 ±0.0 & \textbf{52.5} ±0.2 \\
\cline{1-4}
MOS-10 & 82.6 ±0.0 & 93.7 ±0.0 & \textbf{52.1} ±0.1 \\
\cline{1-4}
MSP & 78.3 ±0.1 & 94.0 ±0.0 & 77.7 ±0.2 \\
\cline{1-4}
PEPR (ours) & 82.5 ±0.9 & 95.8 ±0.3 & 57.4 ±1.0 \\
\cline{1-4}
PEPR-10 (ours) & 83.5 ±0.8 & 96.1 ±0.2 & 57.3 ±0.9 \\
\cline{1-4}
\hline
\end{tabular}


\caption{Summary of key statistics across runs. Mean and standard deviation ($\sigma$) reported across experimental runs. Values are calculated by first taking the mean metric across all datasets for a single experimental run, then calculating the mean and $\sigma$ of the per-run metric across the $10$ trials. Below is the legend for the method names.}
  \begin{itemize}[noitemsep,nolistsep]
    \itemsep0em
    \item CPEPR: see Equation \ref{cpepr}
    \item CPEPR-10: average of Equation \ref{cpepr} via $10$ regressors
    \item EPOW: $S(x) = \psi \frac{1}{n} \sum_i^n (z_i)^2$ 
    \item KLM: KL Matching \cite{hendrycks2019scaling}
    \item MLGT: Max Logit \cite{hendrycks2019scaling}
    \item MOS: Minimum Other Score \cite{huang2021mos}
    \item MOS-10: Average of MOS across $10$ classifiers 
    \item MSP: Maximum Softmax Probability \cite{hendrycks2016baseline}
    \item PEPR: see Equation \ref{pepr}
    \item PEPR-10: average of Equation \ref{pepr} via $10$ regressors
  \end{itemize}

\label{summary_by_run}

\end{table}







\subsection{CPEPR versus Existing Methods:} 
We summarize mean results with variance across runs in Table \ref{summary_by_run}. We compare with competitive methods in the literature that also do not rely on auxiliary outlier data. We include methods tested on large datasets, including MSP \cite{hendrycks2016baseline}, MOS \cite{huang2021mos}, KL matching \cite{hendrycks2019scaling}, and Max Logit \cite{hendrycks2019scaling}. These methods, except for MOS, are trained using a flat, non-grouped, softmax. MOS is trained using grouped softmax, which uses the 8-class groups of \cite{huang2021mos}. 

Desirably, C-PEPR outperforms all other methods in terms of AUROC and AUPR by at least $1$ standard deviation. It does perform worse than MOS in terms of FPR95, but it should be noted that FPR95 measures the false positive rate at an arbitrary recall level. The AUROC curve provides a better representation of the false positive rate at all recall levels. C-PEPR also maintains the same execution speed characteristics as MOS as it only needs to compute a few extra fully-connected layers. 
Using the embedding FNN resulted in a slightly lower validation classification accuracy compared with MOS ($\approx 75.1$ versus $\approx 77.4$). However, this does not seem to negatively affect the OOD detection performance of the model. 

\subsection{On the Bias-Variance Tradeoff:} 
Compared with MOS, PEPR and CPEPR appear to benefit more from a $10$-fold ensemble. PEPR-10 is 1 standard deviation better than MOS-10, but PEPR is not statistically different from MOS, in terms of AUROC. We believe that this is due to the greater standard deviation across all metrics for PEPR and CPEPR. Compared with other OOD methods, our method appears to reduce bias at the expense of increased variance. This is much more noticeable when analyzing results for specific datasets. However, when we consider variance across datasets, PEPR and CPEPR have lower AUROC  (11.0 vs 13.2) and AUPR  (2.7 vs 5.8) standard deviation than MOS . This suggests that the our method performs more consistently across datasets. Overall, this is an advantage that allows users to choose between efficiency and accuracy. 

\subsection{On Issues with Benchmark Selection:} 
We can clearly observe that OOD methods vary greatly in performance across different datasets. For example, in Table \ref{fpr95_table}, KL Matching outperforms MOS by a significant margin for amphibians, fish, and mammals. If \cite{huang2021mos} decided to include these three species benchmarks in their paper, MOS would perform worse than KL Matching in terms of AUROC ($79.6$ versus $80.0$). If we only  evaluated our methods (PEPR \& C-PEPR) using only the four benchmarks provided by \cite{huang2021mos}, then our AUROC performance would not exceed the state-of-the-art ($81.2$ versus $89.6$). As such, evaluation of OOD detection methods may suffer from a definition problem. While researchers agree on the definition of a single out-of-distribution image, they may not agree on the relative weighting of out-of-distribution benchmarks. 

Should a researcher consider anomalous species \cite{hendrycks2019scaling} as one dataset, equal in weight to the Places benchmark? We elected to treat anomalous species as multiple out-of-distribution datasets in order to the follow precedent set by \cite{hendrycks2019scaling}, but we note that this more heavily weights the images in the anomalous species dataset. The choice of out-of-distribution datasets greatly affects the performance of OOD detection methods. This stands in contrast to other machine learning research areas, such as image classification, where improvements in one dataset often correlate with improvements in another. 

The issue of benchmark selection is not only limited to selecting the out-of-distribution datasets. Recent work by \cite{fort2021exploring} showed that they have achieved effectively the ideal AUROC on CIFAR10 versus CIFAR100. It is unclear, however, whether or not their approach would outperform contemporary large-scale OOD methods. In light of the above issues, we encourage further research into OOD benchmark design so that the community may progress towards a consensus on which benchmarks should be used (and the nature in how they are used) for evaluation. 

\subsection{Poor Performance on the Textures Dataset:} 
Detailed results for each dataset are presented in the Table \ref{fpr95_table}. CPEPR and PEPR perform very poorly ($<$60 AUROC) on the textures dataset, which may be explained by the poor performance of the EPOW method. This would suggest that PEPR is limited/hindered by EPOW performance; the usefulness of the estimate of the embeddings is affected by the very embeddings themselves. However, this hypothesis does not apply for the micro-organisms dataset, in which PEPR achieves more than $80$ AUROC and EPOW achieves less than $55$ AUROC. This phenomenon warrants further investigation by future researchers. 

\subsection{On Efficient Evaluation and Reproduction:} 
We compare our results for contemporary methods with those from \cite{huang2021mos} using their four datasets. Our pre-computed backbone outputs setup (see Section \ref{experimentsetup}) differs from previous approaches/setups but reduces training time by $80$\%, which results in a significant reduction in electricity usage and thus green house gas emissions. We see that our reproduction of MOS method achieves an average AUROC across the four datasets (Textures, SUN, Places, iNaturalist) of $89.6$ versus the $90.1$  reported in the original paper \cite{huang2021mos}. We also observe a better average FPR95 of $38.8$ versus $40.0$ on the same four datasets for the MOS method. We believe that our experimental setup faithfully represents MOS and other contemporary methods while significantly reducing the carbon footprint via pre-computed backbone outputs and fewer require training steps. 

\section{Conclusion}
\label{sec:conclusion}

In this work, we proposed a novel OOD detection method, predicted embedding power regression (PEPR), motivated by the properties of intermediate neural layer embeddings. Our experimental results indicate that PEPR performs well for the case of large-scale OOD detection. We train PEPR and other contemporary methods using pre-computed outputs, significantly reducing compute costs. Our experiments across a wide array of datasets shows the PEPR performs better than the state of the art in a statistically significant way. We hope that our study encourages further research in large-scale OOD detection and provides machine learning practitioners with new tools to improve artificial intelligence safety. 

\section*{Acknowledgements}
This material is based upon work supported by the United States National Science Foundation under grant \#2225354

\begin{table*}
\small
\setlength\tabcolsep{4.5pt}

\begin{tabular}{lllllllllll}
\hline
Method & \rotatebox{70}{C-PEPR} & \rotatebox{70}{C-PEPR-10} & \rotatebox{70}{EPOW} & \rotatebox{70}{KLM} & \rotatebox{70}{MLGT} & \rotatebox{70}{MOS} & \rotatebox{70}{MOS-10} & \rotatebox{70}{MSP} & \rotatebox{70}{PEPR} & \rotatebox{70}{PEPR-10} \\
Dataset AUROC &   &  &  &  &  &  &  &  &  &  \\
\hline
Places & 85.9 ±1.9 & 86.1 ±1.9 & 83.7 ±0.8 & 78.4 ±0.1 & 78.2 ±0.1 & \textbf{89.7} ±0.1 & \textbf{89.8} ±0.1 & 78.2 ±0.1 & 83.9 ±1.7 & 84.2 ±1.7 \\
\cline{1-11}
SUN & 87.9 ±2.0 & 88.1 ±2.0 & 84.3 ±0.8 & 81.5 ±0.1 & 78.3 ±0.1 & \textbf{92.5} ±0.1 & \textbf{92.6} ±0.0 & 80.2 ±0.1 & 86.2 ±1.9 & 86.5 ±1.8 \\
\cline{1-11}
Textures & 56.0 ±2.8 & 55.7 ±2.4 & 50.0 ±1.6 & \textbf{83.2} ±0.1 & 68.0 ±0.1 & 78.7 ±0.4 & \textbf{78.9} ±0.1 & 76.9 ±0.1 & 57.8 ±3.2 & 58.1 ±2.4 \\
\cline{1-11}
amphibians & \textbf{72.2} ±4.7 & 72.1 ±4.4 & 61.3 ±2.7 & 72.1 ±0.2 & 68.6 ±0.1 & 59.1 ±0.3 & 59.2 ±0.1 & \textbf{75.0} ±0.1 & 69.9 ±4.8 & 68.5 ±4.9 \\
\cline{1-11}
arachnids & 82.0 ±3.7 & \textbf{83.2} ±2.5 & 76.2 ±1.5 & 69.5 ±0.1 & 59.0 ±0.2 & 67.5 ±0.3 & 67.7 ±0.2 & 76.7 ±0.1 & 78.7 ±5.3 & \textbf{82.1} ±3.4 \\
\cline{1-11}
fish & \textbf{84.1} ±2.4 & \textbf{85.3} ±2.2 & 78.8 ±1.7 & 79.5 ±0.2 & 73.2 ±0.1 & 72.9 ±0.4 & 72.9 ±0.2 & 78.4 ±0.1 & 80.1 ±2.4 & 83.2 ±2.5 \\
\cline{1-11}
fungi & \textbf{96.7} ±0.4 & \textbf{96.7} ±0.3 & 86.6 ±1.5 & 73.4 ±0.2 & 67.0 ±0.2 & 93.1 ±0.1 & 93.4 ±0.1 & 73.4 ±0.2 & 96.1 ±0.5 & 96.0 ±0.4 \\
\cline{1-11}
iNaturalist & 94.8 ±0.4 & 94.8 ±0.3 & 82.5 ±1.8 & 89.9 ±0.1 & 81.5 ±0.1 & \textbf{97.5} ±0.1 & \textbf{97.5} ±0.0 & 86.5 ±0.1 & 94.0 ±0.5 & 94.2 ±0.3 \\
\cline{1-11}
insects & \textbf{82.4} ±3.4 & \textbf{83.3} ±2.8 & 78.7 ±1.4 & 67.7 ±0.1 & 63.5 ±0.1 & 73.1 ±0.3 & 73.3 ±0.1 & 72.5 ±0.1 & 77.2 ±4.4 & 78.5 ±3.6 \\
\cline{1-11}
mammals & 76.1 ±2.5 & 76.2 ±2.8 & 65.6 ±1.2 & 75.4 ±0.1 & 71.5 ±0.1 & 66.9 ±0.2 & 67.2 ±0.1 & \textbf{77.5} ±0.1 & 74.0 ±2.7 & 73.7 ±2.6 \\
\cline{1-11}
microorganisms & 86.4 ±1.5 & 87.0 ±1.3 & 54.1 ±4.1 & 89.0 ±0.6 & 80.9 ±0.6 & \textbf{93.5} ±0.2 & \textbf{93.7} ±0.1 & 81.7 ±0.7 & 87.4 ±1.5 & 88.8 ±1.3 \\
\cline{1-11}
mollusks & 82.5 ±2.8 & \textbf{84.3} ±1.9 & 74.0 ±1.6 & 69.8 ±0.2 & 69.2 ±0.2 & 75.7 ±0.3 & 75.8 ±0.1 & 69.5 ±0.1 & 79.9 ±3.5 & \textbf{84.4} ±1.9 \\
\cline{1-11}
plants & 95.7 ±0.4 & 95.7 ±0.5 & 88.6 ±1.2 & 91.5 ±0.1 & 83.0 ±0.1 & \textbf{98.0} ±0.0 & \textbf{98.0} ±0.0 & 88.5 ±0.1 & 94.6 ±0.6 & 94.7 ±0.7 \\
\cline{1-11}
protozoa & \textbf{96.1} ±0.3 & \textbf{96.2} ±0.2 & 84.7 ±1.8 & 83.7 ±0.1 & 80.8 ±0.1 & 95.7 ±0.1 & 95.9 ±0.0 & 81.5 ±0.1 & 95.5 ±0.4 & 95.7 ±0.3 \\
\cline{1-11}
Dataset AUPR &   &  &  &  &  &  &  &  &  &  \\
\cline{1-11}
Places & 96.2 ±0.6 & 96.3 ±0.5 & 96.0 ±0.2 & 94.3 ±0.0 & 94.9 ±0.0 & \textbf{97.0} ±0.0 & \textbf{97.1} ±0.0 & 94.5 ±0.0 & 95.4 ±0.5 & 95.6 ±0.5 \\
\cline{1-11}
SUN & 96.7 ±0.6 & 96.8 ±0.5 & 95.8 ±0.3 & 95.0 ±0.0 & 95.0 ±0.0 & \textbf{98.0} ±0.0 & \textbf{98.0} ±0.0 & 95.0 ±0.0 & 96.1 ±0.6 & 96.3 ±0.5 \\
\cline{1-11}
Textures & 90.9 ±0.7 & 90.7 ±0.5 & 89.2 ±0.6 & \textbf{97.2} ±0.0 & 95.1 ±0.0 & 95.9 ±0.1 & 96.0 ±0.0 & \textbf{96.3} ±0.0 & 92.0 ±0.8 & 92.2 ±0.5 \\
\cline{1-11}
amphibians & 92.4 ±1.6 & \textbf{92.4} ±1.6 & 86.4 ±1.4 & 90.2 ±0.1 & 89.9 ±0.1 & 83.0 ±0.1 & 83.1 ±0.1 & \textbf{92.6} ±0.0 & 91.9 ±1.6 & 91.5 ±1.6 \\
\cline{1-11}
arachnids & 95.6 ±1.2 & \textbf{96.0} ±0.9 & 92.7 ±0.6 & 90.3 ±0.0 & 87.7 ±0.1 & 88.0 ±0.1 & 88.1 ±0.1 & 93.2 ±0.1 & 94.9 ±1.6 & \textbf{95.8} ±0.9 \\
\cline{1-11}
fish & \textbf{95.9} ±0.7 & \textbf{96.3} ±0.6 & 93.4 ±0.7 & 93.7 ±0.1 & 92.3 ±0.1 & 89.3 ±0.2 & 89.3 ±0.1 & 93.7 ±0.0 & 94.8 ±0.7 & 95.7 ±0.7 \\
\cline{1-11}
fungi & \textbf{99.2} ±0.1 & \textbf{99.2} ±0.1 & 96.2 ±0.5 & 90.8 ±0.1 & 89.5 ±0.1 & 97.4 ±0.1 & 97.5 ±0.0 & 91.2 ±0.1 & 99.0 ±0.1 & 99.0 ±0.1 \\
\cline{1-11}
iNaturalist & 98.9 ±0.1 & 98.9 ±0.1 & 95.8 ±0.5 & 97.2 ±0.0 & 96.0 ±0.0 & \textbf{99.4} ±0.0 & \textbf{99.4} ±0.0 & 96.9 ±0.0 & 98.7 ±0.1 & 98.8 ±0.1 \\
\cline{1-11}
insects & \textbf{95.4} ±1.1 & \textbf{95.7} ±0.8 & 93.5 ±0.5 & 87.0 ±0.1 & 87.4 ±0.1 & 89.0 ±0.1 & 89.0 ±0.0 & 91.1 ±0.1 & 93.8 ±1.5 & 94.3 ±1.1 \\
\cline{1-11}
mammals & \textbf{93.2} ±1.0 & \textbf{93.2} ±1.2 & 87.5 ±0.6 & 91.2 ±0.1 & 90.8 ±0.1 & 86.3 ±0.1 & 86.4 ±0.0 & 92.3 ±0.0 & 92.7 ±1.0 & 92.7 ±0.9 \\
\cline{1-11}
microorganisms & 99.5 ±0.1 & 99.5 ±0.1 & 97.6 ±0.4 & 99.6 ±0.0 & 99.4 ±0.0 & \textbf{99.7} ±0.0 & \textbf{99.7} ±0.0 & 99.3 ±0.0 & 99.5 ±0.1 & 99.6 ±0.1 \\
\cline{1-11}
mollusks & 95.5 ±0.8 & \textbf{96.0} ±0.5 & 91.5 ±0.7 & 89.3 ±0.1 & 90.0 ±0.1 & 90.1 ±0.1 & 90.2 ±0.0 & 89.1 ±0.1 & 94.8 ±1.0 & \textbf{96.2} ±0.5 \\
\cline{1-11}
plants & 98.8 ±0.1 & 98.8 ±0.2 & 96.8 ±0.4 & 97.6 ±0.0 & 95.7 ±0.0 & \textbf{99.5} ±0.0 & \textbf{99.5} ±0.0 & 96.8 ±0.0 & 98.5 ±0.2 & 98.6 ±0.3 \\
\cline{1-11}
protozoa & \textbf{99.0} ±0.1 & \textbf{99.0} ±0.1 & 95.7 ±0.5 & 94.2 ±0.1 & 94.8 ±0.1 & 98.8 ±0.0 & 98.9 ±0.0 & 94.5 ±0.1 & 98.8 ±0.1 & 98.8 ±0.1 \\
\cline{1-11}
Dataset FPR95 &   &  &  &  &  &  &  &  &  &  \\
\cline{1-11}
Places & \textbf{37.0} ±1.1 & \textbf{37.1} ±0.9 & 65.1 ±2.3 & 76.8 ±0.2 & 84.2 ±0.2 & 44.0 ±0.2 & 43.6 ±0.2 & 78.8 ±0.2 & 38.0 ±1.0 & 38.0 ±1.0 \\
\cline{1-11}
SUN & \textbf{32.1} ±1.0 & \textbf{32.2} ±0.8 & 56.2 ±2.0 & 71.5 ±0.2 & 87.2 ±0.1 & 35.5 ±0.4 & 35.1 ±0.2 & 74.5 ±0.3 & 33.1 ±1.0 & 33.0 ±0.9 \\
\cline{1-11}
Textures & 77.3 ±0.5 & 77.4 ±0.4 & 90.8 ±0.7 & 62.4 ±0.4 & 93.7 ±0.2 & \textbf{64.5} ±0.4 & \textbf{64.7} ±0.2 & 78.9 ±0.3 & 76.2 ±0.6 & 76.1 ±0.6 \\
\cline{1-11}
amphibians & 91.1 ±1.1 & 91.2 ±1.0 & 92.8 ±1.2 & \textbf{81.3} ±0.2 & 88.1 ±0.2 & 86.7 ±0.2 & 86.6 ±0.1 & \textbf{83.7} ±0.3 & 91.5 ±0.8 & 91.5 ±0.8 \\
\cline{1-11}
arachnids & 91.0 ±1.6 & 91.2 ±1.3 & \textbf{77.2} ±3.1 & 88.4 ±0.2 & 96.6 ±0.1 & 87.5 ±0.4 & 87.5 ±0.1 & \textbf{87.2} ±0.2 & 92.1 ±1.0 & 92.1 ±0.9 \\
\cline{1-11}
fish & \textbf{79.4} ±2.7 & 79.5 ±2.4 & \textbf{77.2} ±2.6 & 81.9 ±0.3 & 92.6 ±0.1 & 84.3 ±0.5 & 84.2 ±0.2 & 84.4 ±0.3 & 80.7 ±2.2 & 80.7 ±2.0 \\
\cline{1-11}
fungi & \textbf{11.9} ±2.5 & \textbf{11.9} ±2.0 & 64.4 ±4.8 & 84.1 ±0.3 & 94.7 ±0.1 & 27.2 ±0.6 & 26.2 ±0.2 & 87.3 ±0.2 & 14.6 ±3.2 & 14.3 ±2.4 \\
\cline{1-11}
iNaturalist & 27.2 ±2.9 & 27.4 ±2.5 & 70.6 ±4.1 & 45.6 ±0.3 & 88.8 ±0.2 & \textbf{11.2} ±0.3 & \textbf{10.8} ±0.1 & 61.9 ±0.4 & 29.1 ±3.1 & 28.9 ±2.7 \\
\cline{1-11}
insects & 84.8 ±2.4 & 85.0 ±2.0 & 78.9 ±2.7 & 79.0 ±0.2 & 92.2 ±0.1 & \textbf{75.5} ±0.4 & \textbf{75.4} ±0.2 & 84.2 ±0.3 & 86.0 ±1.8 & 86.1 ±1.6 \\
\cline{1-11}
mammals & 85.4 ±1.3 & 85.5 ±1.1 & 88.5 ±1.3 & \textbf{72.1} ±0.2 & 84.9 ±0.1 & 81.9 ±0.2 & 81.7 ±0.1 & \textbf{75.0} ±0.3 & 85.9 ±1.0 & 86.0 ±1.0 \\
\cline{1-11}
microorganisms & 56.5 ±4.4 & 56.9 ±4.3 & 89.5 ±0.7 & 44.8 ±1.1 & 92.8 ±0.3 & \textbf{28.7} ±1.1 & \textbf{27.3} ±0.3 & 70.5 ±1.5 & 48.0 ±3.6 & 48.1 ±3.5 \\
\cline{1-11}
mollusks & 83.7 ±2.4 & 83.9 ±2.1 & 82.9 ±1.9 & 80.1 ±0.2 & 91.1 ±0.1 & \textbf{73.7} ±0.3 & \textbf{73.4} ±0.1 & 85.6 ±0.2 & 84.8 ±1.8 & 84.9 ±1.7 \\
\cline{1-11}
plants & 22.1 ±2.2 & 22.3 ±1.8 & 54.9 ±4.3 & 46.4 ±0.7 & 85.6 ±0.2 & \textbf{9.1} ±0.3 & \textbf{8.6} ±0.1 & 59.9 ±1.0 & 25.5 ±2.8 & 25.3 ±2.3 \\
\cline{1-11}
protozoa & \textbf{16.5} ±1.9 & \textbf{16.6} ±1.6 & 70.1 ±5.0 & 65.3 ±0.6 & 87.8 ±0.1 & 25.1 ±0.7 & 24.2 ±0.3 & 75.2 ±0.4 & 17.8 ±1.8 & 17.6 ±1.5 \\
\cline{1-11}
\hline
\end{tabular}

\caption{AUROC, AUPR, and FPR95 Results for All Datasets expressed as MEAN±$\sigma$. See table \ref{summary_by_run} for method descriptions}
\label{fpr95_table}
    
\end{table*}

\clearpage
\clearpage 

{\small
\bibliographystyle{ieee_fullname}
\bibliography{egbib}
}

\end{document}